\documentclass[
]{ceurart}

\usepackage{listings}
\usepackage{hyperref, xurl}
\usepackage{multirow}

\begin{document}

\copyrightyear{2024}
\copyrightclause{Copyright for this paper by its authors.
  Use permitted under Creative Commons License Attribution 4.0
  International (CC BY 4.0).}

\conference{The First Workshop on Large Language Models for Evaluation in Information Retrieval, 18 July 2024, Washington DC, United States}

\title{The Challenges of Evaluating LLM Applications: An Analysis of Automated, Human, and LLM-Based Approaches}


\author[1]{Bhashithe Abeysinghe}[%
orcid=0009-0006-4107-8615,
email=babeysinghe@air.org,
]
\cormark[1]
\address[1]{American Institutes for Research, Arlington, VA}

\author[1]{Ruhan Circi}[%
email=rcirci@air.org,
]

\cortext[1]{Corresponding author.}

\begin{abstract}
  Chatbots have been an interesting application of natural language generation since its inception. With novel transformer based Generative AI methods, building chatbots have become trivial. Chatbots which are targeted at specific domains for example medicine and psychology are implemented rapidly. This however, should not distract from the need to evaluate the chatbot responses. Especially because the natural language generation community does not entirely agree upon how to effectively evaluate such applications. With this work we discuss the issue further with the increasingly popular LLM based evaluations and how they correlate with human evaluations. Additionally, we introduce a comprehensive factored evaluation mechanism that can be utilized in conjunction with both human and LLM-based evaluations. We present the results of an experimental evaluation conducted using this scheme in one of our chatbot implementations which consumed educational reports, and subsequently compare automated, traditional human evaluation, factored human evaluation, and factored LLM evaluation. Results show that factor based evaluation produces better insights on which aspects need to be improved in LLM applications and further strengthens the argument to use human evaluation in critical spaces where main functionality is not direct retrieval. 
\end{abstract}


\maketitle

\section{Introduction}

The landscape of chatbot development is rapidly evolving, propelled by advancements in Large Language Model (LLM) APIs. While the pace of development is exciting, there is a gap between building an LLM-powered application and building a reliable system with LLMs. This challenge requires carefully considering whether the final product satisfies all requirements and evaluate it to test its alignment with performance and ethical standards. As highlighted by \cite{srivastava_evaluating_2023}, this evaluation process should encompass both a technical assessment and a trust-oriented framework. It is essential to ensure a balance between operational efficiency and responsible usage.

\begin{figure*}[h]
  \centering
  \includegraphics[width=\linewidth]{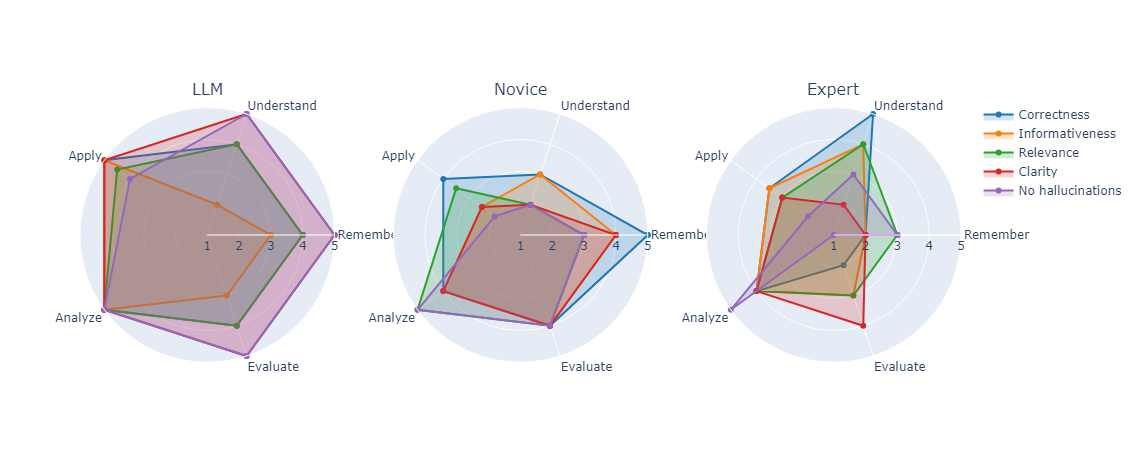}
  \caption{Median of Likert scale ratings of each evaluator. Each spoke shows how an evaluator rated a response based on the question type from Blooms Taxonomy.}\label{factorradar}
\end{figure*}

This process is further complicated by common pitfalls in LLMs, as several authors \cite{gallegos_bias_2023, huang_survey_2023,ji_survey_2023,kaddour_challenges_2023} mention areas of LLM could make mistakes, such as hallucination, tone, and output formatting. Effective evaluation can help to improve and maintain validation and consistency to avoid common pitfalls. The development of an effective evaluation system is timely for researchers and developers alike, given the propagation of LLM based generative applications such as chatbots.

The development cycle of a generic LLM-based application typically covers three phases: a) selection of LLM, b) iterative development of the application, and c) operational deployment of the app. The evaluation of LLMs themselves, as discussed in various papers \cite{guo_evaluating_2023,liang_holistic_2023} is beyond the scope of this brief. However, it is essential to note that the quality of the base LLM is a fundamental component in leveraging its capabilities effectively and minimizing risk in the resulting application. For applications, developers may follow different development approaches (e.g., fine-tuning, chaining, prompting, Retrieval Augmented Generation (RAG), LLM search combined with Knowledge graphs, etc.) and each approach demands tailored evaluation steps e.g., quality of data used in fine-tuning or prompting styles \cite{nori_can_2023}, or chunk size and quantity in RAG \cite{gao_retrieval-augmented_2023}. This paper explores three fundamental approaches for evaluating the final response (i.e., output) generated by LLM-based chatbots namely automated metrics, human evaluation and LLM based evaluation. With respect to human evaluation we investigate preferential evaluation and factored evaluation methods.

\section{Background}
Chatbots interact with users in such a way that they resolve user queries. Some chatbots are domain specific \cite{abd-alrazaq_technical_2020} while others are general purpose chatbots \cite{noauthor_vicuna_2023}. Evaluating a chatbot largely hinges on the intended use and specialization of the chatbot. In reviewing 16 papers on this topic, we summarized several key components that require attention for the evaluation; among these, the clear definition of the chatbot's intended purpose (i.e., use case - that specify business goal or client expectations, and user interaction with app) is critical. Such clarity helps for a focused evaluation of whether the chatbot attains its designated purpose.

The components described in Table 1 suggest that chatbots can be evaluated on different factors (also known as factors or dimensions), such as their ability to answer the users' queries completely, their linguistic effectiveness, and their ability to recall information (either through information retrieval or memory). Additional metrics may include the system's response time, usability, and intuitiveness.

\begin{table}[b]
  \caption{Components of evaluation for an LLM powered application}
  \label{tab:components}
  \begin{tabular}{ll}
    \toprule
    Key components&Description\\
    \midrule
    Response properties  & Correctness of output\\
    & Readability/tone \\
    Grading approach & One uttarance\\
    & Conversation\\
    & Comparative/preference\\
    User experience & Number of interactions per user\\
    & Helpful suggestions from the bot\\
    &The intuitiveness of the application\\
  \bottomrule
\end{tabular}
\end{table}

Currently, there are no common methods or agreed upon best practices that are robust enough to evaluate LLM-based applications. As pointed out in almost all the prior work on this topic, a notable challenge is the lack of consensus on appropriate evaluation criteria and metrics. Therefore, researchers and developers bear the responsibility of choosing evaluation methods that are most appropriate for their unique application. This responsibility may not only increase development timelines but may also lead to underpowered statistical evaluations \cite{card_little_2020,van_der_lee_best_2019}. A resounding issue of automated metrics is that they are inconsistent with results and may not always correlate with human evaluation. But many still prefer to use them in evaluation due to being readily available and also easily repeatable \cite{banerjee_meteor_2005, lin_rouge_2004-1, papineni_bleu_2002, sellam_bleurt_2020,zhang_bertscore_2020}. Which is not the case with human evaluation, it is expensive and will not be repeatable in the same context even if one uses the same humans \cite{finch_dont_2023,van_der_lee_best_2019,van_der_lee_human_2021}. We must acknowledge the work where generative AI models which are being used at the evaluation step such as ChatEval, GPTScore and ARES \cite{chan_chateval_2023,fu_gptscore_2023, saad-falcon_ares_2024} which are novel applications of LLMs. \cite{svikhnushina_approximating_2023} discusses about “bot-play” where an already evaluated LLM being used in evaluating a new un-evaluated LLM. When considering LLM based evaluators, one must make sure the evaluator LLM produces acceptable and accurate decisions to a given threshold.

Human evaluation remains the most widely accepted form of evaluation in research studies despite frequent reports of underpowered results \cite{clark_all_2021,van_der_lee_best_2019}. Several attempts have been called for the standardization of human evaluation methods \cite{howcroft_what_2021,van_der_lee_human_2021}, but its costly nature often leads researchers to report on systems with statistically insufficient power. Additionally, the sensitivity of human evaluators to the framing of questions (framed negatively or positively) is reported to influence outcomes \cite{schoch_this_2020}. For conversational or dialogue systems, the common standard of human evaluation is Quality on Likert scales. Quality can vary across tasks, and it encompasses multiple factors such as correctness, relevance, informativeness, consistency, understanding, etc. \cite{finch_dont_2023}. \cite{van_der_lee_best_2019} suggest using a minimum of 100 questions rated on 5 or 7-point Likert scales to evaluate multiple dimensions.  This seems to be a difficult goal to achieve due to the expensive nature of human evaluation.

The variability in expert opinions has led to multiple recommendations for refining human evaluation approaches. Engaging at least four experts is recommended, but more is preferable for robust results \cite{van_der_lee_human_2021}. However, using expert evaluations may not always be productive, particularly if the system is not designed for expert use \cite{clark_all_2021}. In cases where the number of available experts is limited, a comparative (also known as preferential) evaluation approach is often preferred. Additionally, it is advisable to involve about 10 to 60 non-expert users - the intended end-users of the system - in the evaluation process and to ensure that the Inter Annotator Agreement (IAA) is reported for reliability (refer to Table 3 in \cite{van_der_lee_best_2019} for best practices). It is also imperative to use external evaluators who have not taken part in the conversation to judge the conversation \cite{finch_dont_2023}. \cite{vijayaraghavan_algorithm_2020}  discusses the complexities in explaining human evaluations; noting that individuals with varying levels of expertise can provide divergent assessments of the same response, this again shows the importance of employing many humans with varying expertise to completely evaluate such a system.

In summarizing insights from reviewed research articles, it is evident that human evaluation remains a common and indispensable element in the evaluation pipeline of chatbot systems, albeit implemented at different stages. Additionally, a diverse selection of metrics is frequently employed to assess various aspects of chatbot responses. Utilizing evaluator LLMs seems to be a promising approach that warrants exploration due to its potential to offer efficient and scalable evaluation. While the current focus is on the evaluation, a potentially critical factor, often overlooked, is the nature of the data used for testing and evaluation and many papers lack specificity regarding the types of questions posed to chatbots. We propose that incorporating a range of question types, informed by cognitive psychology frameworks such as Bloom's Taxonomy, could significantly enhance the systematic evaluation of chatbot responses and the insights drawn from such an evaluation. 

\begin{figure}[b]
  \centering
  \includegraphics[width=8cm]{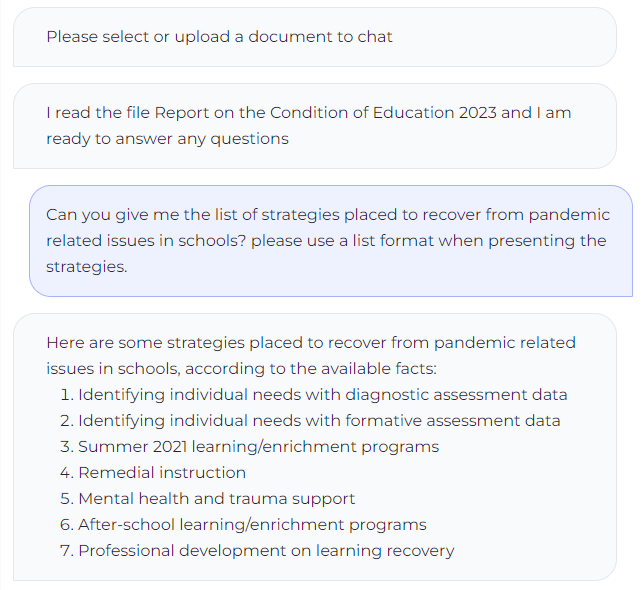}
  \caption{Screen capture of the EdTalk chatbot answering a question}\label{ui}
\end{figure}

To experiment with the evaluation procedures, we implement a chatbot first (Figure \ref{ui}). This implementation follows industry standards such as Retrieval Augmented Generation (RAG), Vector Databases etc. to create a chatbot. The chatbot EdTalk aims to assist users in \emph{navigating and comprehending lengthy reports} by harnessing the power of LLMs and the goals are to have \emph{minimal hallucination} and \emph{strict adherence} to factual information from its knowledge base. The goal of this chatbot is to make the educational reports such as Condition of Education accessible to a wide range of readers. Hence, chatbots knowledge base is built with the said reports. By evaluating EdTalk, we investigate if this chatbot aligns with its initial goals. Simultaneously we find if the chatbot is able to consistently follow the goals for various different types of questions in Bloom's Taxonomy. Later we compare the results from various evaluation procedures including automated, human and LLM-based to find what is more informative with respect to the development of this chatbot.

\begin{table*}[t]
  \caption{Scenario from Condition of Education report 2023, the example question, Annotator expert response, generated response. Similar response pairs are used in the BLEURT evaluation}\label{anotresponses}
  \begin{tabular}{p{0.15\textwidth}p{0.30\textwidth}p{0.52\textwidth}}
    \toprule
          Question &  Expert response  &  Generated response \\
    \midrule 
    What challenges did schools face during the pandemic? & Schools experienced many other challenges—besides staffing-related ones—during their recovery from the coronavirus pandemic. Some of these other issues facing public schools included reported increases in student and teacher absenteeism, student socioemotional and behavioral development, and an increase in the percentage of students seeking mental health services from school, as compared with before the coronavirus pandemic.  & Schools faced many challenges during the pandemic besides staffing-related issues. Some of these challenges included reported increases in student and teacher absenteeism, student socioemotional and behavioral development, and an increase in the percentage of students seeking mental health services from school, as compared with before the coronavirus pandemic. In particular, in April 2022, some 70 percent of public schools reported that the percentage of students who had sought mental health services from school had increased since the start of the coronavirus pandemic. Overall, only 12 percent of schools strongly agreed and 44 percent moderately agreed that their school was able to effectively provide mental health services to all students in need. So, schools faced challenges related to student mental health, absenteeism, and socioemotional and behavioral development during the pandemic. \\
    \bottomrule
  \end{tabular}
\end{table*}

\section{Evaluation procedures}
We understand that chatbots, like any software will have an iterative implementation where the developers would be updating components which make up the chatbot. Each of these components and the full system need to be evaluated for reliability and performance. In this section we dive into various evaluation procedures we conducted and briefly explain how they were implemented. But we only focus on the utterance-based evaluation; meaning that we shall only be investigating procedures which are built to look at responses of the chatbot. Other components performance such as the semantic search used for retrieval in RAG is not in scope for this investigation.

To conduct the evaluation we employ the service of 5 humans. Initially, one of the human evaluators, having access to the content to be evaluated, generated 40 questions based on Bloom's Taxonomy \cite{armstrong_blooms_2010}. The purpose behind adopting Bloom's Taxonomy was to determine the efficacy of the chatbot in responding to different types of questions. This approach adds another unique dimension to the evaluation process, enabling us to evaluate the quality of the chatbot's responses against different types of questions. It should be noted that the specific questions used in the evaluation were dependent on the use case of the chatbot implementation and have not been disclosed in this article.

Then a pair of humans hereafter known as - \emph{annotators}, write their own responses to the above questions. Later another pair hereafter known as - \emph{evaluators} determines the quality of the responses. Both pairs consists of an expert and a novice. An expert is someone who has been working with these reports for more than 2 years and a novice is new to the area but has some experience with the content.

\subsection{Automated evaluation}
Selecting an automated evaluation model is one crucial step. We do not select n-gram based methods because of the issues that literature points out and hence, we utilize embedding based methods. In that regard we believe BLUERT \cite{sellam_bleurt_2020} to be the best out of the selection. We must not forget that this methods would still sometimes produce inconsistent results, but as it is repeatable, it can be used at the rapid development stage to test parameters such as chunk sizes, overlap sizes etc. BLEURT requires a reference text and a generated text to compute similarity, and \cite{sellam_bleurt_2020} suggests using a specific checkpoint to achieve best comparison\footnote{\url{https://github.com/google-research/bleurt?tab=readme-ov-file\#checkpoints}}, an example of the reference text (Expert response) and the generated text (generated response) is given in the Table \ref{anotresponses}. Evaluating if the chatbot responses are similar to \emph{annotators} is straightforward with BLEURT.

\subsection{Human evaluation}
Human evaluation on the other hand is a bit complex. There is traditional human evaluation which is typically a preferential rating of what response a human would prefer more. While this is an acceptable measure \cite{van_der_lee_best_2019}, it may still miss insights from the results. We conduct this traditional preferential evaluation first to start the human evaluation. The humans do not need to be experts in the domain to conduct this type of evaluation \cite{clark_all_2021}.

Then we enlist evaluators to rate responses of the chatbot for the previously created questions. Rating will be conducted on a few factors \cite{fu_gptscore_2023,van_der_lee_best_2019}. We carefully select these factors so that we can effectively evaluate many aspects of the chatbot, where many of the selected factors were inspired by \cite{van_der_lee_best_2019}. We develop a 5-point Likert scale-based questionnaire from which we collect expert ratings for the chatbot responses. 

Instructions on how to perform the ratings were given prior to the \emph{evaluators}. Table \ref{tab:criteria} shows what questions an evaluator should ask before rating a response for a criterion. The criterions are set up so that a response with all the accurate and relevant information, without unnecessary information, in the most clear and concise manner is rated high. We also take hallucinations into the equation as well; this covers most quality criteria a generative AI application should look for. Evaluators are also free to refer the text where the questions re based off of, but we did not make the previous Annotator responses available for the Evaluators. We gave example ratings for a few questions and responses which were not part of the 40 selected above, these included examples for ratings 1, 3 and 5. Evaluators were free to determine how to assign the intermediate ratings. 

\begin{table}[b]
  \caption{Criteria for the Likert scale questionnaire}
  \label{tab:criteria}
  \begin{tabular}{p{0.20\textwidth}p{0.50\textwidth}}
    \toprule
    Criterion&Description\\
    \midrule
    Relevance&If the facts presented are required by the question?\\
    Informativeness&Are all the facts called by the question presented by the response?\\
    Correctness&How correct the generated response?\\
    Clarity&Does the question call for a certain formatting ofr the answer or is the response brief or verbose?\\
    hallucination&Is the answer a hallucinated reference, information etc.?\\
  \bottomrule
\end{tabular}
\end{table}

\subsection{LLM-based evaluation}
The evaluation procedure being discussed is a relatively new one, and there is currently limited literature available to support its reliability as compared to human evaluation. The purpose of this study is to contribute to the existing literature by comparing human-based evaluation with LLM-based evaluation. The researchers used the same instructions that were given to human evaluators to prompt the LLM for evaluation. In addition, examples for each Likert scale value were provided to ensure that the LLM was aligned with the evaluation criteria, this is the only difference between the human instructions as humans do not receive examples for all Likert scales. The evaluation prompt included the question, facts retrieved from the content, and the response generated by the chatbot, as per the methodology proposed by \cite{saad-falcon_ares_2024}. The responses were evaluated for a given factor at a time, and the generated evaluation responses were processed to extract similar Likert scales from the LLM. The LLM evaluators did not have access to the Annotator responses created in the automated evaluation step, but LLM evaluator did have access to the content of the document. This allowed the researchers to compare the LLM-based evaluation with the human evaluation in a similar light.

\section{Results}
In this section, the results of all evaluation procedures are compared and contrasted. The purpose is to gain an understanding of what was learned from each experiment and to identify any advantages or disadvantages associated with each method. Bloom's Taxonomy is used to make comparisons, but the specific types within the taxonomy are not explained in this work.

\begin{table}[t]
  \caption{Automated evaluation results; each generated answer is compared against a human (Expert or Novice) and the BLEURT score is reported herewith}\label{autoresults}
  \begin{tabular}{lrrr}
    \toprule
          Type &  Expert  &    Novice \\
    \midrule 
    Remember   &     0.45 &      0.40 \\   
    Understand &     0.61 &      0.55 \\     
    Apply      &     0.44 &      0.24 \\  
    Analyze    &     0.47 &      0.41 \\  
    Evaluate   &     0.22 &      0.31 \\
    \bottomrule
  \end{tabular}
\end{table}

Table \ref{autoresults} presents the results captured by the automated evaluation experiment. As we explain in the previous sections, here we use BLEURT \cite{sellam_bleurt_2020} as the metric to compute similarities of the generated response against a human written answer. This evaluation can be conducted rapidly if the human written responses are readily available. Meaning that the human needs to only write the response once, where it is possible to repeatedly run the evaluation after the parameters of the application are altered. It is not clear how to compare two BLEURT scores for a similar task where multiple reference text are used. Upon inspection and comparison of BLEURT values, it was noted that for some question types, expert and novice fell into similar ranges. For both humans, the generated response has a lower similarity in \emph{Evaluate} questions. For \emph{Apply} questions, while Experts similarity is at 0.44, novice has 0.24. Highest similarities were reported in both humans at \emph{Understand} questions.


\begin{table}[b]
  \caption{Percentage of preference of generated response in the preferential rating evaluation}\label{prefresults}
  \begin{tabular}{lr}
    \toprule
          Type &  Generated response preference \\
    \midrule
      Remember &   31\% \\
    Understand &   100\% \\
         Apply &   0\% \\
       Analyze &   57\% \\
      Evaluate &   33\% \\
    \bottomrule
    \end{tabular}
    
\end{table}

\begin{table}[t!]
  \centering
  \caption{Agreement of human annotators using Krippendorff's alpha}\label{krip}
  \begin{tabular}{lr}
    \toprule
      Criterion & Krippendorff's $\alpha$ \\
    \midrule              
      Correctness & 0.12 \\
      Informativeness & 0.18 \\
      Relevance   & 0.31 \\
      Clarity     & 0.52 \\
      Hallucinations & 0.31\\            
    \bottomrule
    \end{tabular}    
\end{table}

\begin{table*}[b]
  \caption{Factored evaluation results; median across question type. Higher the better.}\label{facresults}
  \begin{tabular}{clccccc}
    \toprule
    &    Type &    Correctness &  Informativeness &  Relevance   &  Clarity   &  Hallucinations  \\
    \midrule
    \multirow{5}{*}{Expert} & Remember   &   \textbf{2} &  \textbf{2}      &   \textbf{3} & \textbf{2} &    \textbf{3} \\
                            & Understand &   \textbf{5} &          4       &            4 &          2 &             3 \\ 
                            & Apply      &   \textbf{3.5} &        3.5     &      3       &          3 &       2       \\
                            & Analyze    &   \textbf{4} &          4       &            4 &          4 &             5 \\
                            & Evaluate   &   \textbf{2} &          3       &            3 &          4 &             1 \\ \hline
    \multirow{5}{*}{Novice} & Remember   &   \textbf{5} &  \textbf{4}      &   \textbf{3} & \textbf{4} &    \textbf{3} \\
                            & Understand &   \textbf{3} &          3       &            2 &          2 &             2 \\ 
                            & Apply      &   \textbf{4} &          2.5     &          3.5 &        2.5 &             2 \\
                            & Analyze    &   \textbf{4} &          4       &            5 &          4 &             5 \\
                            & Evaluate   &   \textbf{4} &          4       &            4 &          4 &             4 \\ \hline
    \multirow{5}{*}{LLM}    & Remember   &   \textbf{4} &  \textbf{3}      &   \textbf{4} & \textbf{5} &    \textbf{5} \\
                            & Understand &   \textbf{4} &          2       &            4 &          5 &             5 \\ 
                            & Apply      &   \textbf{5} &          5       &      4.5     &          5 &             4 \\
                            & Analyze    &   \textbf{5} &          5       &            5 &          5 &             5 \\
                            & Evaluate   &   \textbf{4} &          3       &            4 &          5 &             5 \\
    \bottomrule
    \end{tabular}
\end{table*}

We conducted traditional human evaluation through preferential rating first, this type of evaluation does not require domain experts to conduct evaluation and is much faster considering the other human evaluation methods. Here we find that the chatbots answers are preferred only 47\% (on average) of the time, Table \ref{prefresults} present results broken down into the same Bloom's Taxonomy type. This measure does not reveal anything about what areas are needed improvement in order to perform better. Which is typically why the community prefers factored human evaluation.

Table \ref{facresults} reports the results of the factored evaluation in both human and LLM procedures. Since we used Likert scales to capture ratings, we have reported the results via medians of each factor and question type. The visualized results are displayed in Figure \ref{factorradar}, which clearly highlight the notable differences between novices and experts in their approaches to response analysis. The graph underscores the importance of recognizing individual variations in cognitive processing and interpretation of information.

Using the factored human evaluation procedure, we were able to experimentally figure out previously elusive facts about the generative application. When we initially conducted trivial automated and human evaluation (preferential), if we do not break questions down to Bloom's Taxonomy, we only get one measure to test if the chatbot works within the parameters of an acceptable application. This is not usually enough to understand the underlying complex issues of LLMs, and if they are present in the LLM-powered application or not. RAG systems are built to retrieve information which is available in context. This means that when posed with \emph{Remember} questions, they must perform well, but as the results from the expert show; EdTalk does not perform well with \emph{Remember} questions (Table \ref{facresults} and Figure \ref{factorradar}). It shows also that chatbot responses are not consistent enough to say anything related to other question types. This result reveals while RAG chatbots should be great at answering retrieval based questions they sometimes do not work as intended in the perspective of a human. We also note that the automated evaluation with BLEURT showed similar patterns with each of the question type as well, but when we take the novice into account, the similarity is not present anymore. One advantage in this type of evaluation is that we can now check the inter-rater reliability, and we show this in Table \ref{krip}. We notice the major issue pointed out by many prior work here with, where humans not agreeing in their reviews. Also by categorizing questions into factors we notice that human agreement is moderate in \emph{Clarity} but all other factors are low agreement. One disadvantage we notice here is the ability of repeating the evaluation effort, same humans may rate these responses differently if we change the order or the framing of the questions in the questionnaire \cite{van_der_lee_best_2019,clark_all_2021}.

\section{Discussion}
The goal of this work is to illustrate how challenging it is to evaluate an LLM based application, especially evaluating a chatbot with current methodologies including automated, human and LLM procedures. We first demonstrate that there are advantages and disadvantages in all three of these approaches. We also note the differences of results gained from all three evaluation procedures, there is very little correlation between these results and it would be difficult to suggest one to be used. We also observed that the experts evaluation results are a bit stricter and resulted lower scores generally for many factors. The novice had looked at the chatbot in a favorable light and we notice the slightly elevated scores. Using an LLM to evaluate the chatbot responses seems to be not reliable as the LLM scores its own responses high. In our experimental case, we used the same LLM (GPT-3.5) to generate the responses and also as the evaluator LLM. This is not the ideal setting as \cite{svikhnushina_approximating_2023} points out, in \cite{svikhnushina_approximating_2023} authors point out if an LLM is not evaluated it must be evaluated using an already evaluated LLM or a higher order LLM. Given this situation of uncertain evaluations from any procedure, we should not distract the readers from the need for evaluating. To improve the reliability of evaluation, we suggest increasing the number of humans used in the factored human evaluation. Also enlisting a wide range of expertise would create a smoothed preview of the results; however, this would increase the expensiveness of the evaluation. As \cite{van_der_lee_best_2019} suggests, enlisting a larger amount of intended users of a chatbot would still not be ideal as these users may also create confusion on whats correct and whats not. Allowing untrained humans to make judgments on the factors will not yield the most accurate results, similar to the case we have with LLM results in Figure \ref{factorradar}.

One deciding factor would be the repeatability and the amount of funds a person has toward evaluating a chatbot. In this regard we note while automated procedures are repeatable, low reliability of these metrics make a case against them. Human evaluation is considered the gold standard, while that can be true research indicates that the human disagreement is a greater issue; we also notice this issue indicated in Table \ref{krip}. LLM evaluators are a novel adaptation of LLMs, its greatest adversary right now is not having enough research to support its reliability. We observe that in some cases LLM evaluators have similar responses to human evaluators. But this is not the case always, in most instances LLM evaluators tend to be overly confident in the response being correct. We cannot reject the promise in LLM evaluators as we can set various personalities and take various versions of its evaluation rapidly \cite{chan_chateval_2023}, but this also must be explored in terms of whether a person of such an expertise would rate the same response in a similar way. Further research needs to be conducted in understanding how LLMs can help us evaluate LLMs.

\begin{acknowledgments}
  Abhinav Cheruvu for helping with implementation of the chatbot and to Tabitha Tezil, Erika Kessler and Jijun Zhang for helping with human evaluation. 
\end{acknowledgments}

\bibliography{refs}

\begin{thebibliography}{29}
\expandafter\ifx\csname natexlab\endcsname\relax\def\natexlab#1{#1}\fi
\providecommand{\url}[1]{\texttt{#1}}
\providecommand{\href}[2]{#2}
\providecommand{\path}[1]{#1}
\providecommand{\DOIprefix}{doi:}
\providecommand{\ArXivprefix}{arXiv:}
\providecommand{\URLprefix}{URL: }
\providecommand{\Pubmedprefix}{pmid:}
\providecommand{\doi}[1]{\href{http://dx.doi.org/#1}{\path{#1}}}
\providecommand{\Pubmed}[1]{\href{pmid:#1}{\path{#1}}}
\providecommand{\bibinfo}[2]{#2}
\ifx\xfnm\relax \def\xfnm[#1]{\unskip,\space#1}\fi
\bibitem[{Srivastava et~al.(2023)Srivastava, Lakkaraju, Koppel, Narayanan,
  Kundu, and Joshi}]{srivastava_evaluating_2023}
\bibinfo{author}{B.~Srivastava}, \bibinfo{author}{K.~Lakkaraju},
  \bibinfo{author}{T.~Koppel}, \bibinfo{author}{V.~Narayanan},
  \bibinfo{author}{A.~Kundu}, \bibinfo{author}{S.~Joshi},
  \bibinfo{title}{Evaluating {Chatbots} to {Promote} {Users}' {Trust} --
  {Practices} and {Open} {Problems}}, \bibinfo{year}{2023}. \URLprefix
  \url{http://arxiv.org/abs/2309.05680}, \bibinfo{note}{arXiv:2309.05680 [cs]}.
\bibitem[{Gallegos et~al.(2023)Gallegos, Rossi, Barrow, Tanjim, Kim,
  Dernoncourt, Yu, Zhang, and Ahmed}]{gallegos_bias_2023}
\bibinfo{author}{I.~O. Gallegos}, \bibinfo{author}{R.~A. Rossi},
  \bibinfo{author}{J.~Barrow}, \bibinfo{author}{M.~M. Tanjim},
  \bibinfo{author}{S.~Kim}, \bibinfo{author}{F.~Dernoncourt},
  \bibinfo{author}{T.~Yu}, \bibinfo{author}{R.~Zhang}, \bibinfo{author}{N.~K.
  Ahmed}, \bibinfo{title}{Bias and {Fairness} in {Large} {Language} {Models}:
  {A} {Survey}}, \bibinfo{year}{2023}. \URLprefix
  \url{http://arxiv.org/abs/2309.00770}.
  \DOIprefix\doi{10.48550/arXiv.2309.00770}, \bibinfo{note}{arXiv:2309.00770
  [cs]}.
\bibitem[{Huang et~al.(2023)Huang, Yu, Ma, Zhong, Feng, Wang, Chen, Peng, Feng,
  Qin, and Liu}]{huang_survey_2023}
\bibinfo{author}{L.~Huang}, \bibinfo{author}{W.~Yu}, \bibinfo{author}{W.~Ma},
  \bibinfo{author}{W.~Zhong}, \bibinfo{author}{Z.~Feng},
  \bibinfo{author}{H.~Wang}, \bibinfo{author}{Q.~Chen},
  \bibinfo{author}{W.~Peng}, \bibinfo{author}{X.~Feng},
  \bibinfo{author}{B.~Qin}, \bibinfo{author}{T.~Liu}, \bibinfo{title}{A
  {Survey} on {Hallucination} in {Large} {Language} {Models}: {Principles},
  {Taxonomy}, {Challenges}, and {Open} {Questions}}, \bibinfo{year}{2023}.
  \URLprefix \url{http://arxiv.org/abs/2311.05232}.
  \DOIprefix\doi{10.48550/arXiv.2311.05232}, \bibinfo{note}{arXiv:2311.05232
  [cs]}.
\bibitem[{Ji et~al.(2023)Ji, Lee, Frieske, Yu, Su, Xu, Ishii, Bang, Madotto,
  and Fung}]{ji_survey_2023}
\bibinfo{author}{Z.~Ji}, \bibinfo{author}{N.~Lee},
  \bibinfo{author}{R.~Frieske}, \bibinfo{author}{T.~Yu},
  \bibinfo{author}{D.~Su}, \bibinfo{author}{Y.~Xu}, \bibinfo{author}{E.~Ishii},
  \bibinfo{author}{Y.~J. Bang}, \bibinfo{author}{A.~Madotto},
  \bibinfo{author}{P.~Fung},
\newblock \bibinfo{title}{Survey of {Hallucination} in {Natural} {Language}
  {Generation}},
\newblock \bibinfo{journal}{ACM Computing Surveys} \bibinfo{volume}{55}
  (\bibinfo{year}{2023}) \bibinfo{pages}{1--38}. \URLprefix
  \url{https://dl.acm.org/doi/10.1145/3571730}.
  \DOIprefix\doi{10.1145/3571730}.
\bibitem[{Kaddour et~al.(2023)Kaddour, Harris, Mozes, Bradley, Raileanu, and
  McHardy}]{kaddour_challenges_2023}
\bibinfo{author}{J.~Kaddour}, \bibinfo{author}{J.~Harris},
  \bibinfo{author}{M.~Mozes}, \bibinfo{author}{H.~Bradley},
  \bibinfo{author}{R.~Raileanu}, \bibinfo{author}{R.~McHardy},
  \bibinfo{title}{Challenges and {Applications} of {Large} {Language}
  {Models}}, \bibinfo{year}{2023}. \URLprefix
  \url{http://arxiv.org/abs/2307.10169}.
  \DOIprefix\doi{10.48550/arXiv.2307.10169}, \bibinfo{note}{arXiv:2307.10169
  [cs]}.
\bibitem[{Guo et~al.(2023)Guo, Jin, Liu, Huang, Shi, Supryadi, Yu, Liu, Li,
  Xiong, and Xiong}]{guo_evaluating_2023}
\bibinfo{author}{Z.~Guo}, \bibinfo{author}{R.~Jin}, \bibinfo{author}{C.~Liu},
  \bibinfo{author}{Y.~Huang}, \bibinfo{author}{D.~Shi},
  \bibinfo{author}{Supryadi}, \bibinfo{author}{L.~Yu},
  \bibinfo{author}{Y.~Liu}, \bibinfo{author}{J.~Li},
  \bibinfo{author}{B.~Xiong}, \bibinfo{author}{D.~Xiong},
  \bibinfo{title}{Evaluating {Large} {Language} {Models}: {A} {Comprehensive}
  {Survey}}, \bibinfo{year}{2023}. \URLprefix
  \url{http://arxiv.org/abs/2310.19736}.
  \DOIprefix\doi{10.48550/arXiv.2310.19736}, \bibinfo{note}{arXiv:2310.19736
  [cs]}.
\bibitem[{Liang et~al.(2023)Liang, Bommasani, Lee, Tsipras, Soylu, Yasunaga,
  Zhang, Narayanan, Wu, Kumar, Newman, Yuan, Yan, Zhang, Cosgrove, Manning,
  Ré, Acosta-Navas, Hudson, Zelikman, Durmus, Ladhak, Rong, Ren, Yao, Wang,
  Santhanam, Orr, Zheng, Yuksekgonul, Suzgun, Kim, Guha, Chatterji, Khattab,
  Henderson, Huang, Chi, Xie, Santurkar, Ganguli, Hashimoto, Icard, Zhang,
  Chaudhary, Wang, Li, Mai, Zhang, and Koreeda}]{liang_holistic_2023}
\bibinfo{author}{P.~Liang}, \bibinfo{author}{R.~Bommasani},
  \bibinfo{author}{T.~Lee}, \bibinfo{author}{D.~Tsipras},
  \bibinfo{author}{D.~Soylu}, \bibinfo{author}{M.~Yasunaga},
  \bibinfo{author}{Y.~Zhang}, \bibinfo{author}{D.~Narayanan},
  \bibinfo{author}{Y.~Wu}, \bibinfo{author}{A.~Kumar},
  \bibinfo{author}{B.~Newman}, \bibinfo{author}{B.~Yuan},
  \bibinfo{author}{B.~Yan}, \bibinfo{author}{C.~Zhang},
  \bibinfo{author}{C.~Cosgrove}, \bibinfo{author}{C.~D. Manning},
  \bibinfo{author}{C.~Ré}, \bibinfo{author}{D.~Acosta-Navas},
  \bibinfo{author}{D.~A. Hudson}, \bibinfo{author}{E.~Zelikman},
  \bibinfo{author}{E.~Durmus}, \bibinfo{author}{F.~Ladhak},
  \bibinfo{author}{F.~Rong}, \bibinfo{author}{H.~Ren},
  \bibinfo{author}{H.~Yao}, \bibinfo{author}{J.~Wang},
  \bibinfo{author}{K.~Santhanam}, \bibinfo{author}{L.~Orr},
  \bibinfo{author}{L.~Zheng}, \bibinfo{author}{M.~Yuksekgonul},
  \bibinfo{author}{M.~Suzgun}, \bibinfo{author}{N.~Kim},
  \bibinfo{author}{N.~Guha}, \bibinfo{author}{N.~Chatterji},
  \bibinfo{author}{O.~Khattab}, \bibinfo{author}{P.~Henderson},
  \bibinfo{author}{Q.~Huang}, \bibinfo{author}{R.~Chi}, \bibinfo{author}{S.~M.
  Xie}, \bibinfo{author}{S.~Santurkar}, \bibinfo{author}{S.~Ganguli},
  \bibinfo{author}{T.~Hashimoto}, \bibinfo{author}{T.~Icard},
  \bibinfo{author}{T.~Zhang}, \bibinfo{author}{V.~Chaudhary},
  \bibinfo{author}{W.~Wang}, \bibinfo{author}{X.~Li}, \bibinfo{author}{Y.~Mai},
  \bibinfo{author}{Y.~Zhang}, \bibinfo{author}{Y.~Koreeda},
  \bibinfo{title}{Holistic {Evaluation} of {Language} {Models}},
  \bibinfo{year}{2023}. \URLprefix \url{http://arxiv.org/abs/2211.09110}.
  \DOIprefix\doi{10.48550/arXiv.2211.09110}, \bibinfo{note}{arXiv:2211.09110
  [cs]}.
\bibitem[{Nori et~al.(2023)Nori, Lee, Zhang, Carignan, Edgar, Fusi, King,
  Larson, Li, Liu, Luo, McKinney, Ness, Poon, Qin, Usuyama, White, and
  Horvitz}]{nori_can_2023}
\bibinfo{author}{H.~Nori}, \bibinfo{author}{Y.~T. Lee},
  \bibinfo{author}{S.~Zhang}, \bibinfo{author}{D.~Carignan},
  \bibinfo{author}{R.~Edgar}, \bibinfo{author}{N.~Fusi},
  \bibinfo{author}{N.~King}, \bibinfo{author}{J.~Larson},
  \bibinfo{author}{Y.~Li}, \bibinfo{author}{W.~Liu}, \bibinfo{author}{R.~Luo},
  \bibinfo{author}{S.~M. McKinney}, \bibinfo{author}{R.~O. Ness},
  \bibinfo{author}{H.~Poon}, \bibinfo{author}{T.~Qin},
  \bibinfo{author}{N.~Usuyama}, \bibinfo{author}{C.~White},
  \bibinfo{author}{E.~Horvitz}, \bibinfo{title}{Can {Generalist} {Foundation}
  {Models} {Outcompete} {Special}-{Purpose} {Tuning}? {Case} {Study} in
  {Medicine}}, \bibinfo{year}{2023}. \URLprefix
  \url{http://arxiv.org/abs/2311.16452}.
  \DOIprefix\doi{10.48550/arXiv.2311.16452}, \bibinfo{note}{arXiv:2311.16452
  [cs]}.
\bibitem[{Gao et~al.(2023)Gao, Xiong, Gao, Jia, Pan, Bi, Dai, Sun, and
  Wang}]{gao_retrieval-augmented_2023}
\bibinfo{author}{Y.~Gao}, \bibinfo{author}{Y.~Xiong}, \bibinfo{author}{X.~Gao},
  \bibinfo{author}{K.~Jia}, \bibinfo{author}{J.~Pan}, \bibinfo{author}{Y.~Bi},
  \bibinfo{author}{Y.~Dai}, \bibinfo{author}{J.~Sun},
  \bibinfo{author}{H.~Wang}, \bibinfo{title}{Retrieval-{Augmented} {Generation}
  for {Large} {Language} {Models}: {A} {Survey}}, \bibinfo{year}{2023}.
  \URLprefix \url{http://arxiv.org/abs/2312.10997}.
  \DOIprefix\doi{10.48550/arXiv.2312.10997}, \bibinfo{note}{arXiv:2312.10997
  [cs]}.
\bibitem[{Abd-Alrazaq et~al.(2020)Abd-Alrazaq, Safi, Alajlani, Warren, Househ,
  and Denecke}]{abd-alrazaq_technical_2020}
\bibinfo{author}{A.~Abd-Alrazaq}, \bibinfo{author}{Z.~Safi},
  \bibinfo{author}{M.~Alajlani}, \bibinfo{author}{J.~Warren},
  \bibinfo{author}{M.~Househ}, \bibinfo{author}{K.~Denecke},
\newblock \bibinfo{title}{Technical {Metrics} {Used} to {Evaluate} {Health}
  {Care} {Chatbots}: {Scoping} {Review}},
\newblock \bibinfo{journal}{Journal of Medical Internet Research}
  \bibinfo{volume}{22} (\bibinfo{year}{2020}) \bibinfo{pages}{e18301}.
  \URLprefix \url{http://www.jmir.org/2020/6/e18301/}.
  \DOIprefix\doi{10.2196/18301}.
\bibitem[{noa(2023)}]{noauthor_vicuna_2023}
\bibinfo{title}{Vicuna: {An} {Open}-{Source} {Chatbot} {Impressing} {GPT}-4
  with 90\%* {ChatGPT} {Quality} {\textbar} {LMSYS} {Org}},
  \bibinfo{year}{2023}. \URLprefix
  \url{https://lmsys.org/blog/2023-03-30-vicuna}.
\bibitem[{Card et~al.(2020)Card, Henderson, Khandelwal, Jia, Mahowald, and
  Jurafsky}]{card_little_2020}
\bibinfo{author}{D.~Card}, \bibinfo{author}{P.~Henderson},
  \bibinfo{author}{U.~Khandelwal}, \bibinfo{author}{R.~Jia},
  \bibinfo{author}{K.~Mahowald}, \bibinfo{author}{D.~Jurafsky},
  \bibinfo{title}{With {Little} {Power} {Comes} {Great} {Responsibility}},
  \bibinfo{year}{2020}. \URLprefix \url{http://arxiv.org/abs/2010.06595},
  \bibinfo{note}{arXiv:2010.06595 [cs]}.
\bibitem[{van~der Lee et~al.(2019)van~der Lee, Gatt, Van~Miltenburg, Wubben,
  and Krahmer}]{van_der_lee_best_2019}
\bibinfo{author}{C.~van~der Lee}, \bibinfo{author}{A.~Gatt},
  \bibinfo{author}{E.~Van~Miltenburg}, \bibinfo{author}{S.~Wubben},
  \bibinfo{author}{E.~Krahmer},
\newblock \bibinfo{title}{Best practices for the human evaluation of
  automatically generated text},
\newblock in: \bibinfo{booktitle}{Proceedings of the 12th {International}
  {Conference} on {Natural} {Language} {Generation}}, \bibinfo{year}{2019}, pp.
  \bibinfo{pages}{355--368}.
\bibitem[{Banerjee and Lavie(2005)}]{banerjee_meteor_2005}
\bibinfo{author}{S.~Banerjee}, \bibinfo{author}{A.~Lavie},
\newblock \bibinfo{title}{{METEOR}: {An} {Automatic} {Metric} for {MT}
  {Evaluation} with {Improved} {Correlation} with {Human} {Judgments}},
\newblock in: \bibinfo{editor}{J.~Goldstein}, \bibinfo{editor}{A.~Lavie},
  \bibinfo{editor}{C.-Y. Lin}, \bibinfo{editor}{C.~Voss} (Eds.),
  \bibinfo{booktitle}{Proceedings of the {ACL} {Workshop} on {Intrinsic} and
  {Extrinsic} {Evaluation} {Measures} for {Machine} {Translation} and/or
  {Summarization}}, \bibinfo{publisher}{Association for Computational
  Linguistics}, \bibinfo{address}{Ann Arbor, Michigan}, \bibinfo{year}{2005},
  pp. \bibinfo{pages}{65--72}. \URLprefix
  \url{https://aclanthology.org/W05-0909}.
\bibitem[{Lin(2004)}]{lin_rouge_2004-1}
\bibinfo{author}{C.-Y. Lin},
\newblock \bibinfo{title}{{ROUGE}: {A} {Package} for {Automatic} {Evaluation}
  of {Summaries}},
\newblock in: \bibinfo{booktitle}{Text {Summarization} {Branches} {Out}},
  \bibinfo{publisher}{Association for Computational Linguistics},
  \bibinfo{address}{Barcelona, Spain}, \bibinfo{year}{2004}, pp.
  \bibinfo{pages}{74--81}. \URLprefix \url{https://aclanthology.org/W04-1013}.
\bibitem[{Papineni et~al.(2002)Papineni, Roukos, Ward, and
  Zhu}]{papineni_bleu_2002}
\bibinfo{author}{K.~Papineni}, \bibinfo{author}{S.~Roukos},
  \bibinfo{author}{T.~Ward}, \bibinfo{author}{W.-J. Zhu},
\newblock \bibinfo{title}{Bleu: a {Method} for {Automatic} {Evaluation} of
  {Machine} {Translation}},
\newblock in: \bibinfo{editor}{P.~Isabelle}, \bibinfo{editor}{E.~Charniak},
  \bibinfo{editor}{D.~Lin} (Eds.), \bibinfo{booktitle}{Proceedings of the 40th
  {Annual} {Meeting} of the {Association} for {Computational} {Linguistics}},
  \bibinfo{publisher}{Association for Computational Linguistics},
  \bibinfo{address}{Philadelphia, Pennsylvania, USA}, \bibinfo{year}{2002}, pp.
  \bibinfo{pages}{311--318}. \URLprefix
  \url{https://aclanthology.org/P02-1040}.
  \DOIprefix\doi{10.3115/1073083.1073135}.
\bibitem[{Sellam et~al.(2020)Sellam, Das, and Parikh}]{sellam_bleurt_2020}
\bibinfo{author}{T.~Sellam}, \bibinfo{author}{D.~Das}, \bibinfo{author}{A.~P.
  Parikh}, \bibinfo{title}{{BLEURT}: {Learning} {Robust} {Metrics} for {Text}
  {Generation}}, \bibinfo{year}{2020}. \URLprefix
  \url{http://arxiv.org/abs/2004.04696}.
  \DOIprefix\doi{10.48550/arXiv.2004.04696}, \bibinfo{note}{arXiv:2004.04696
  [cs]}.
\bibitem[{Zhang et~al.(2020)Zhang, Kishore, Wu, Weinberger, and
  Artzi}]{zhang_bertscore_2020}
\bibinfo{author}{T.~Zhang}, \bibinfo{author}{V.~Kishore},
  \bibinfo{author}{F.~Wu}, \bibinfo{author}{K.~Q. Weinberger},
  \bibinfo{author}{Y.~Artzi}, \bibinfo{title}{{BERTScore}: {Evaluating} {Text}
  {Generation} with {BERT}}, \bibinfo{year}{2020}. \URLprefix
  \url{http://arxiv.org/abs/1904.09675}.
  \DOIprefix\doi{10.48550/arXiv.1904.09675}, \bibinfo{note}{arXiv:1904.09675
  [cs]}.
\bibitem[{Finch et~al.(2023)Finch, Finch, and Choi}]{finch_dont_2023}
\bibinfo{author}{S.~E. Finch}, \bibinfo{author}{J.~D. Finch},
  \bibinfo{author}{J.~D. Choi}, \bibinfo{title}{Don't {Forget} {Your} {ABC}'s:
  {Evaluating} the {State}-of-the-{Art} in {Chat}-{Oriented} {Dialogue}
  {Systems}}, \bibinfo{year}{2023}. \URLprefix
  \url{http://arxiv.org/abs/2212.09180}, \bibinfo{note}{arXiv:2212.09180 [cs]}.
\bibitem[{van~der Lee et~al.(2021)van~der Lee, Gatt, van Miltenburg, and
  Krahmer}]{van_der_lee_human_2021}
\bibinfo{author}{C.~van~der Lee}, \bibinfo{author}{A.~Gatt},
  \bibinfo{author}{E.~van Miltenburg}, \bibinfo{author}{E.~Krahmer},
\newblock \bibinfo{title}{Human evaluation of automatically generated text:
  {Current} trends and best practice guidelines},
\newblock \bibinfo{journal}{Computer Speech \& Language} \bibinfo{volume}{67}
  (\bibinfo{year}{2021}) \bibinfo{pages}{101151}. \URLprefix
  \url{https://www.sciencedirect.com/science/article/pii/S088523082030084X}.
  \DOIprefix\doi{10.1016/j.csl.2020.101151}.
\bibitem[{Chan et~al.(2023)Chan, Chen, Su, Yu, Xue, Zhang, Fu, and
  Liu}]{chan_chateval_2023}
\bibinfo{author}{C.-M. Chan}, \bibinfo{author}{W.~Chen},
  \bibinfo{author}{Y.~Su}, \bibinfo{author}{J.~Yu}, \bibinfo{author}{W.~Xue},
  \bibinfo{author}{S.~Zhang}, \bibinfo{author}{J.~Fu},
  \bibinfo{author}{Z.~Liu}, \bibinfo{title}{{ChatEval}: {Towards} {Better}
  {LLM}-based {Evaluators} through {Multi}-{Agent} {Debate}},
  \bibinfo{year}{2023}. \URLprefix \url{http://arxiv.org/abs/2308.07201},
  \bibinfo{note}{arXiv:2308.07201 [cs]}.
\bibitem[{Fu et~al.(2023)Fu, Ng, Jiang, and Liu}]{fu_gptscore_2023}
\bibinfo{author}{J.~Fu}, \bibinfo{author}{S.-K. Ng},
  \bibinfo{author}{Z.~Jiang}, \bibinfo{author}{P.~Liu},
  \bibinfo{title}{{GPTScore}: {Evaluate} as {You} {Desire}},
  \bibinfo{year}{2023}. \URLprefix \url{http://arxiv.org/abs/2302.04166},
  \bibinfo{note}{arXiv:2302.04166 [cs]}.
\bibitem[{Saad-Falcon et~al.(2024)Saad-Falcon, Khattab, Potts, and
  Zaharia}]{saad-falcon_ares_2024}
\bibinfo{author}{J.~Saad-Falcon}, \bibinfo{author}{O.~Khattab},
  \bibinfo{author}{C.~Potts}, \bibinfo{author}{M.~Zaharia},
  \bibinfo{title}{{ARES}: {An} {Automated} {Evaluation} {Framework} for
  {Retrieval}-{Augmented} {Generation} {Systems}}, \bibinfo{year}{2024}.
  \URLprefix \url{http://arxiv.org/abs/2311.09476},
  \bibinfo{note}{arXiv:2311.09476 [cs]}.
\bibitem[{Svikhnushina and Pu(2023)}]{svikhnushina_approximating_2023}
\bibinfo{author}{E.~Svikhnushina}, \bibinfo{author}{P.~Pu},
\newblock \bibinfo{title}{Approximating {Online} {Human} {Evaluation} of
  {Social} {Chatbots} with {Prompting}},
\newblock in: \bibinfo{editor}{S.~Stoyanchev}, \bibinfo{editor}{S.~Joty},
  \bibinfo{editor}{D.~Schlangen}, \bibinfo{editor}{O.~Dusek},
  \bibinfo{editor}{C.~Kennington}, \bibinfo{editor}{M.~Alikhani} (Eds.),
  \bibinfo{booktitle}{Proceedings of the 24th {Meeting} of the {Special}
  {Interest} {Group} on {Discourse} and {Dialogue}},
  \bibinfo{publisher}{Association for Computational Linguistics},
  \bibinfo{address}{Prague, Czechia}, \bibinfo{year}{2023}, pp.
  \bibinfo{pages}{268--281}. \URLprefix
  \url{https://aclanthology.org/2023.sigdial-1.25}.
\bibitem[{Clark et~al.(2021)Clark, August, Serrano, Haduong, Gururangan, and
  Smith}]{clark_all_2021}
\bibinfo{author}{E.~Clark}, \bibinfo{author}{T.~August},
  \bibinfo{author}{S.~Serrano}, \bibinfo{author}{N.~Haduong},
  \bibinfo{author}{S.~Gururangan}, \bibinfo{author}{N.~A. Smith},
  \bibinfo{title}{All {That}'s '{Human}' {Is} {Not} {Gold}: {Evaluating}
  {Human} {Evaluation} of {Generated} {Text}}, \bibinfo{year}{2021}. \URLprefix
  \url{http://arxiv.org/abs/2107.00061}, \bibinfo{note}{arXiv:2107.00061 [cs]}.
\bibitem[{Howcroft and Rieser(2021)}]{howcroft_what_2021}
\bibinfo{author}{D.~M. Howcroft}, \bibinfo{author}{V.~Rieser},
\newblock \bibinfo{title}{What happens if you treat ordinal ratings as interval
  data? {Human} evaluations in {NLP} are even more under-powered than you
  think},
\newblock in: \bibinfo{editor}{M.-F. Moens}, \bibinfo{editor}{X.~Huang},
  \bibinfo{editor}{L.~Specia}, \bibinfo{editor}{S.~W.-t. Yih} (Eds.),
  \bibinfo{booktitle}{Proceedings of the 2021 {Conference} on {Empirical}
  {Methods} in {Natural} {Language} {Processing}},
  \bibinfo{publisher}{Association for Computational Linguistics},
  \bibinfo{address}{Online and Punta Cana, Dominican Republic},
  \bibinfo{year}{2021}, pp. \bibinfo{pages}{8932--8939}. \URLprefix
  \url{https://aclanthology.org/2021.emnlp-main.703}.
  \DOIprefix\doi{10.18653/v1/2021.emnlp-main.703}.
\bibitem[{Schoch et~al.(2020)Schoch, Yang, and Ji}]{schoch_this_2020}
\bibinfo{author}{S.~Schoch}, \bibinfo{author}{D.~Yang},
  \bibinfo{author}{Y.~Ji},
\newblock \bibinfo{title}{“{This} is a {Problem}, {Don}'t {You} {Agree}?”
  {Framing} and {Bias} in {Human} {Evaluation} for {Natural} {Language}
  {Generation}},
\newblock in: \bibinfo{editor}{S.~Agarwal}, \bibinfo{editor}{O.~Dušek},
  \bibinfo{editor}{S.~Gehrmann}, \bibinfo{editor}{D.~Gkatzia},
  \bibinfo{editor}{I.~Konstas}, \bibinfo{editor}{E.~Van~Miltenburg},
  \bibinfo{editor}{S.~Santhanam} (Eds.), \bibinfo{booktitle}{Proceedings of the
  1st {Workshop} on {Evaluating} {NLG} {Evaluation}},
  \bibinfo{publisher}{Association for Computational Linguistics},
  \bibinfo{address}{Online (Dublin, Ireland)}, \bibinfo{year}{2020}, pp.
  \bibinfo{pages}{10--16}. \URLprefix
  \url{https://aclanthology.org/2020.evalnlgeval-1.2}.
\bibitem[{Vijayaraghavan et~al.(2020)Vijayaraghavan, Cooper, and
  J.}]{vijayaraghavan_algorithm_2020}
\bibinfo{author}{V.~Vijayaraghavan}, \bibinfo{author}{J.~B. Cooper},
  \bibinfo{author}{R.~L. J.},
\newblock \bibinfo{title}{Algorithm {Inspection} for {Chatbot} {Performance}
  {Evaluation}},
\newblock \bibinfo{journal}{Procedia Computer Science} \bibinfo{volume}{171}
  (\bibinfo{year}{2020}) \bibinfo{pages}{2267--2274}. \URLprefix
  \url{https://linkinghub.elsevier.com/retrieve/pii/S1877050920312370}.
  \DOIprefix\doi{10.1016/j.procs.2020.04.245}.
\bibitem[{Armstrong(2010)}]{armstrong_blooms_2010}
\bibinfo{author}{P.~Armstrong}, \bibinfo{title}{Bloom’s {Taxonomy}},
  \bibinfo{year}{2010}. \URLprefix
  \url{https://cft.vanderbilt.edu/guides-sub-pages/blooms-taxonomy/}.

\end{thebibliography}

\appendix
\section{Prompts}
This section notes the prompts that have been used in this work, we first note the prompt that has been utilized in the RAG process in the chatbot for clarity and then a sample prompt that was

\subsection{RAG Prompt}
\begin{lstlisting}[breaklines]
The user asks the question <question>. Here are some facts that could be used to support the question, <facts delimited by semicolons>.
      
You must first investigate if it is possible to support an answer with the available facts If you do not have facts to support an answer, step by step explaining your reasoning behind each action you must come up with a answer by processing, applying and evaluating facts as needed. Otherwise you must only respond with "I dont know" and do not output anything else.
\end{lstlisting}

\subsection{LLM Evaluator Prompt}
Here in this prompt we only add the prompt used with the ``Correctness'' criterion and similar prompts can be drawn for others.
\begin{lstlisting}[breaklines]
You are an expert education researcher.
You are given a set of facts, a question that relates to the text of these facts and an answer for the given question. 
Your task is to evaluate if the answer is a good answer to the given question based off of a criterion and also considering the facts. 
Evaluation steps: 
1. Read the facts: Start by carefully reading the facts provided. Understand the context, main points, and any relevant details. 
2. Analyze the Question: Examine the question that relates to the facts. Ensure you have a clear understanding of what the question is asking for. 
3. Review the Answer: Carefully read the answer provided and assess it based only on the following criterion: 

Correctness: Does the answer provide accurate information based on the paragraph text? 

4. Assign a Score: Use the 5-point scale to assign a score to the answer: 
Score 1: If the answer is wrong compared with the facts for the question
Score 2: If the answer is mostly wrong compared with the facts for the question
Score 3: If the answer is partly correct given the facts for the question
Score 4: If the answer is mostly correct given the facts for the question
Score 5: If the answer is correct given the facts for the question

5. Document Scores: Keep a record of the scores and feedback for reference. 
This can be helpful for tracking progress and ensuring consistency in your evaluations. 

The facts: \verb|<relevant_facts>|

The Question for this facts: \verb|<question>|

The Answer: \verb|<response>|

Score: 
\end{lstlisting}
\end{document}